\documentclass[letterpaper, 10 pt, journal, twoside]{IEEEtran}
\usepackage{amsmath,amsfonts}
\usepackage{algorithmic}
\usepackage{algorithm}
\usepackage{array}
\usepackage[caption=false,font=normalsize,labelfont=sf,textfont=sf]{subfig}
\usepackage{textcomp}
\usepackage{stfloats}
\usepackage{url}
\usepackage{verbatim}
\usepackage{graphicx}
\usepackage{cite}
\usepackage[symbol]{footmisc}
\usepackage[flushleft]{threeparttable}
\usepackage[table]{xcolor}
\usepackage{multirow}
\usepackage{enumitem}

\begin{document}

\title{Optimized Kalman Filter based State Estimation and Height Control in Hopping Robots}

\author{Samuel Burns$^{1}$ and Matthew Woodward$^{1}$
\thanks{Manuscript received: August, 21, 2024; Revised June, 4, 2025.}
\thanks{$^{1} $The authors are with the Robot Locomotion and Biomechanics Laboratory, Mechanical Engineering Department, Tufts University, Medford, MA, 01890
        {\tt\footnotesize matthew.woodward@tufts.edu}}%
}

\markboth{arXiv. Preprint Version. Submitted June, 2025.}%
{Burns \MakeLowercase{\textit{et al.}}: State Estimation}



\maketitle

\begin{abstract}
Rotor-based hopping locomotion significantly improves efficiency and operation time as compared to purely flying systems; where most hopping robots use the liftoff states and an assumed ballistic trajectory to determine the hopping height. However, significant aerial phase force (e.g., thrust and drag) can invalidate this assumption and lead to poor estimation performance. To combat this issue, a group has implemented multiple sensors (active and passive optical, inertial, and contact) and significant computational power to achieve full state estimation. This, however, poses a significant challenge to the development of light-weight, high-performance, low observable, jamming and electronic interference resistant hopping systems; especially in perceptually degraded environments (e.g., dust, smoke). Here we show a training procedure for a coupled hopping phase and Kalman filter-based vertical state estimator, requiring only inertial measurements, which is able to learn the characteristics of the target system, sensors, locomotion behaviors, environment, and acceleration measurement aliasing conditions. The resulting estimator, given hop heights up to 4 m and velocities up to $\pm7$ m/s, achieves a mean absolute percent error in the hop apex height of 12.5\% with an aerial trajectory average normalized mean absolute error in position and velocity of 19\% and 16.5\%, respectively; while operating at 840 Hz, on a dual-core 240 MHz processor, with a total robot mass of 672 g. Due to the low mass and computational power, the presented estimator could also be used as a degraded operational mode in cases of sensor damage, malfunction, or occlusion in more complex robots.
\end{abstract}


\section{Introduction}
\IEEEPARstart{H}{opping} robots combine the agility of aerial systems with the efficiency of terrestrial systems \cite{burns_design_2025,bai_agile_2024,wang_terrestrial_2023}; where the focus has been placed on predicting the vertical states for control, from the liftoff characteristics \cite{Raibert1984,Raibert1984a}. However, this method is not applicable to the recent generation of rotor-based hopping robots that may not follow strictly ballistic trajectories in the aerial phase \cite{burns_design_2025, Zhu2022, wang_terrestrial_2023, kang_fast_2024,bai_agile_2024}. Therefore, a new method for continuous state estimation is necessary. This is particularly challenging in light-weight hopping robots, without the payload for significant increases in sensors, processing, and power; furthermore high-performance hopping robots, where aliasing of sensors measurements can be significant, especially during the impacts associated with hopping. The challenge is further compounded in perceptually degraded environments (e.g., dust, smoke, darkness), in contested environments in which active jamming may be present, and where sensing modalities that radiate energy into the environment may lead to detection and destruction. Currently, robot rotational state estimation can be achieved through inertial measurement units (IMUs), and translational state estimation can be achieved through GPS and combined optical flow and range finder senors; however, range finder sensors radiate energy into the environment for sensing, and can have difficultly operating in perceptually degraded environments. As IMUs are ubiquitous in robots and small in size, this work seeks to develop an IMU-based vertical state estimation methodology for hopping height control that could be extended to horizontal states in the future; fully replacing the optical flow sensor for operation in highly perceptually degraded environments.

To date numerous non-continuous jumping and hopping robots have been developed including miniature jumping robots \cite{Kovac2008g, zaitsev_locust-inspired_2015, shin_towards_2008, Churaman2011, Noh2012a, churaman_first_2012, Koh2013f}, high specific energy robots \cite{Hawkes2022,BostonDynamics2013}, soft jumping robots \cite{Tolley2014, bartlett_3d-printed_2015}, multimodal robots \cite{pan_tumro_2024}, with wings for gliding \cite{Woodward2011, kovac_epfl_2011, Woodward2014, desbiens_efficient_2013, desbiens_design_2014, beck_jump_2017, Woodward2018, Woodward2018a, woodward_tailored_2019}, and inertial tails or reaction wheels for orientation control \cite{Zhao2015b,haldane_repetitive_2017, yim_precision_2018, yim_drift-free_2019, Yim2020}. These robots have also used a wide variety of hopping leg designs including: linkage \cite{li_jumping_2009, Zhao2013, Plecnik2017, fiorini_development_2003}, and linear \cite{zhao_development_2009, scarfogliero_design_2007, ho_novel_2012, aguilar_robophysical_2016} legs. These robots represent a highly diverse range of characteristics and constraints. However, given the minimal impact of an IMU integration, the addition of the proposed method is highly feasible for any system.

Currently, five untethered continuous hopping robots exist: MultiMo-MHR (our system) \cite{burns_design_2025}, PogoDrone \cite{Zhu2022}, Hopcopter \cite{bai_agile_2024}, Salto/Salto-1P \cite{Haldane2016,haldane_power_2016,Plecnik2017,lee_self-engaging_2018, haldane_repetitive_2017, yim_precision_2018, yim_drift-free_2019,Yim2020}, and PogoX \cite{wang_terrestrial_2023,kang_fast_2024} and one tethered continuous insect-scale hopping robot \cite{hsiao_hybrid_2025}. While PogoDrone and the insect-scale hopping robot shows no form of state estimation, both the Hopcopter and Salto robots use the estimated or measured liftoff states and an assumed ballistic aerial trajectory to estimate the hopping height. However, the ballistic trajectory assumption invalids this technique for robots with applied aerial phase forces. PogoX (robot mass = 3.65 kg) displays the most thorough state estimation to date, using an extended Kalman filter, moving horizon estimation, and visual inertial odometry, requiring multiple sensors, advanced computational hardware, large weight budget, and direct measurement of both vertical position and acceleration to estimate vertical velocity; thus posing a challenge to broad applicability. This work seeks to estimate both position and velocity using only direct acceleration measurements allowing for a broader integration across platforms while also being used in a degraded operational mode when primary sensors are damaged or malfunction in more complex systems; such as PogoX.

This paper is organized as follows. Section 1 introduces the topic, and Section 2 briefly covers the robot design with focus on the changes from previous work \cite{burns_design_2025}. Section 3 discusses the hopping phase estimator, and Section 4 presents the hopping state estimator with associated training procedure and results. Section 5 explores the effects of the sensing frequency on measurement aliasing and ultimately state estimation. Section 6 presents the experiential procedure and results of the deployed state estimation technique on the MultiMo-MHR, and Section 7 summarizes the work.

\begin{figure*}[tbp]
\centerline{\includegraphics[width=1.0\textwidth]{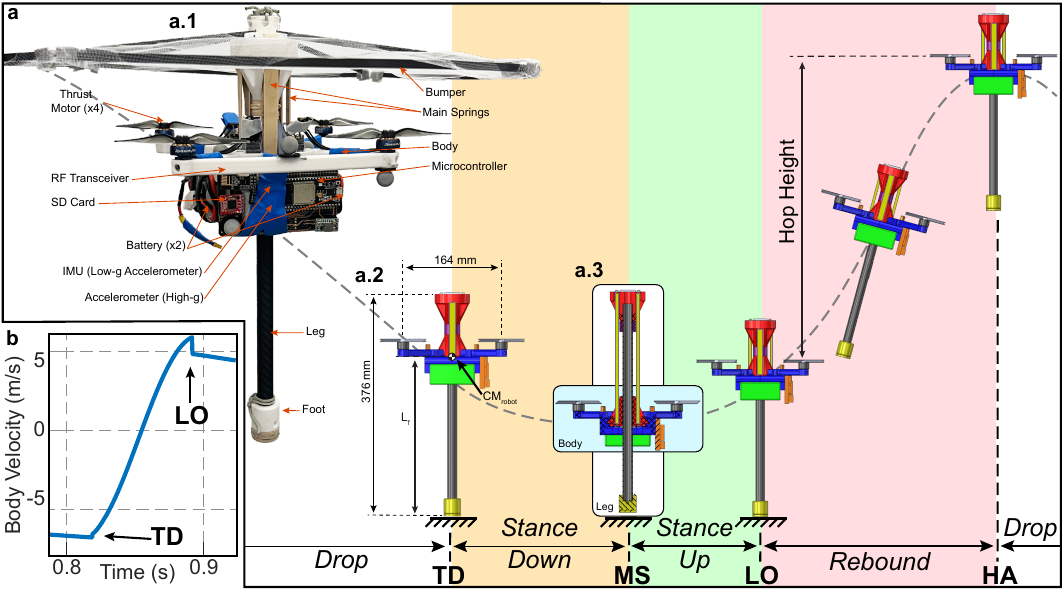}}
\caption{Hopping robot. a) Operational phases (drop, stance down, stance up, rebound) and transitions (touchdown (TD), max squat (MS), liftoff (LO), hop apex (HA)) a.1) Photo of the MultiMo-MHR platform with components labeled. a.2) CAD model with overall dimensions labeled. a.3) CAD model section view with spring extension (stance phases) showing the combined body and leg structures. b) Dynamic simulation showing the locally, highly non-linear, behavior of the body at TD and LO for a 3 m hop.}
\label{fig:robot}
\end{figure*}

\section{Robot Design}
The MultiMo-MHR, developed in previous work \cite{burns_design_2025}, is a monopedal hopping robot that utilizes rotors for orientation control and energy input (Fig. \ref{fig:robot}a, Appendix Robot Parameters). The robot consists of two bodies where a single unactuated prismatic joint, with the motion-axis aligned to the leg-axis, connects the body to the leg. At touchdown the body's momentum causes it to move down the leg, extending the main power springs. The springs consist of multiple rubber bands that are able to store a maximum usable locomotion energy of approximately 34 J resulting in a maximum hopping height of over 5 meters; where additional bands can increase the height.

The MultiMo-MHR uses an on-board microcontroller (ESP32-WROOM-32, two Tensilica Xtensa LX6 cores at 240MHz) for real-time control (Fig. \ref{fig:robot}a). Given the significant difference in the accelerations experienced during ground contact and the aerial phases, the platform has two inertial measurement units (IMUs), including a low-g (MPU-6050: $\pm 16$g, $\pm$2000 deg/s, 1000 Hz) and high-g (H3LIS331DL: $\pm 100$g, 1000Hz), ensuring accurate measurements tailored to the varying magnitudes of forces encountered; where the static measurement noises are $\pm 0.013g$ (MPU-6050) and $\pm 0.32g$ (H3LIS331DL). A data logger (Sparkfun OpenLog) is used to store both the sensor data and calculated values during operation. The robot is powered by a 14.7v battery pack (two 7.4v, 2s, 2200 mAh, 35C batteries in series) to provide sufficient power to the four electronic speed controllers (ESC, 20 Amp) and motors (SpeedyBee 1404 4500 kv). Finally, a radio frequency transceiver (nRF24L01) receives data from an external PC which is used to simulate additional sensors, via a motion capture system (Vicon vantage V5), while focusing on hopping height control (e.g. PMW3901 Optical Flow Sensor [horizontal positions] and  WT901 [orientation]).

\begin{figure}[tbp]
\centerline{\includegraphics[width=0.5\textwidth]{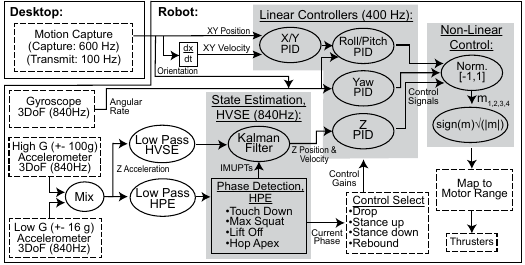}}
\caption{Hopping robot control block diagram.}
\label{fig:ctrl}
\end{figure}

The previously developed controller, composed of the horizontal [X,Y], orientation [roll,pitch,yaw], and locomotion energy (LE) input [Z or height] control, is augmented with a vertical state estimator and a new LE input controller for hopping height control (Fig. \ref{fig:ctrl}). Due to the highly damped nature of the vertical states, a proportional controller is used for the LE input control. The P-controller output is mapped to a thrust-to-weight ratio (TWR) of between 0\% and 83.7\% (i.e., 100\% equates to the same thrust required for hover); where the minimum motor command is set to 5\% duty-cycle to increase the motor's responsiveness while producing almost no thrust. The controller operates as a sliding mode control at ranges greater than 0.33 m from the desired height.

\begin{figure}[tbp]
\centerline{\includegraphics[width=0.45\textwidth]{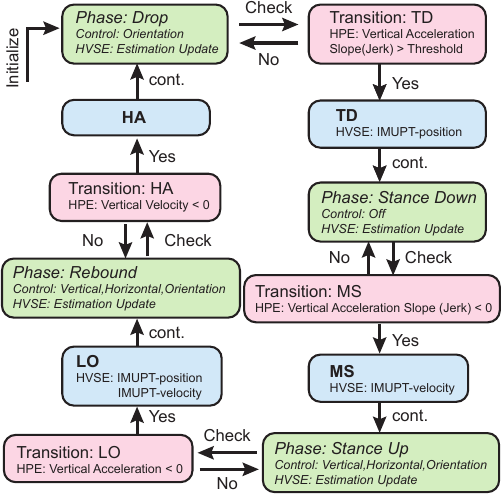}}
\caption{State machine diagram of the Hopping Phase Estimator (HPE), Hopping Vertical State Estimator (HVSE), and controllers showing both the prediction updates and Inferred Measurement Updates (IMUPTs) of the HVSE. Shown are phases (green), phase transition criteria (red), and transitions (blue)}
\label{fig:stateMachine}
\end{figure}

\section{Hopping Phase Estimator (HPE)}
A hopping phase estimator (HPE) is developed (Fig. \ref{fig:stateMachine}) where the operational phases (green), phase transition criteria (red), and transitions (blue) are shown. The HPE determines the \textit{drop}, \textit{stance down}, \textit{stance up}, and \textit{rebound} phases through identification of the four transitions; touchdown (TD), maximum squat (MS), liftoff (LO), and hop apex (HA). Previous work in hopping robots has indicated the TD and LO transitions can be determined from the states \cite{Raibert1984}, with two specific methods including leg motion through the use of a leg encoder \cite{Haldane2016,haldane_repetitive_2017,yim_precision_2018,yim_drift-free_2019,Yim2020} and ground contact through a force sensor on the foot \cite{wang_terrestrial_2023,kang_fast_2024}. However, both methods require an additional sensor with complex mounting requirements beyond the typical 9-axis inertial measurement unit (IMU) required for aerial phase control. Instead, inspired by work in inertial pedestrian navigation or pedestrian dead-reckoning \cite{wang_stance-phase_2015,norrdine_step_2016,ju_pedestrian_2016,foxlin_pedestrian_2005}, multiple acceleration measurements (IMU) are used to calculate the slope (jerk) and a set positive threshold determines TD. This same measurement is then used to determine MS as the point at which the jerk becomes negative. LO is determined by a single filtered acceleration measurement that is less than zero; indicating gravity is the dominant acceleration. Finally, previous works have predicted the HA from measured liftoff states and an assumed ballistic trajectory. However, as hopping height increases so to will drag, resulting in increased deviation from gravity-based ballistic motion. Furthermore, rotor-based hopping robots can significantly alter their aerial trajectory from that of ballistic motion; due to thrust produced in the aerial phase. While range finding sensors can directly measure the height and thus the HA, they significantly increase the observability of the system, provide and opening for external interference, and increased height can degrade the measurement. As proprioceptive sensors are unable to directly or indirectly detect it, the transition criteria for the HA requires an accurate estimation of the vertical velocity.

\begin{table*}[tbp!]
\caption{Kalman Filters}
\begin{center}
\begin{tabular}{p{0.05\textwidth}|p{0.24\textwidth}|p{0.315\textwidth}|p{0.19\textwidth}|p{0.085\textwidth}}
\hline
\textbf{Filter} & \textbf{Filter Equations} & \textbf{Filter Parameters} & \textbf{IMUPTs} & \textbf{Errors} \\
\textbf{Type} & \textbf{} & \textbf{} & \textbf{(Phase/Transition)} & (998 hops) \\
\hline
& KF - Prediction & $z_k = H \hat{x}_k + n_a, \, u_k = [0,0,1] \, a_k^{w}$ & \\
KF1 & {\begin{tabular}{l}
    $x_k^- = F x_{k-1} + G u_k$\\
    $P_k^- = F P_{k-1} F^T + Q$\\
\end{tabular}} & 
    $F = \begin{bmatrix}
        1 & \Delta t \\
        0 & 1
    \end{bmatrix}
    , \,
    G = 
    \begin{bmatrix}
        0.5 \Delta t^2 \\
        \Delta t
    \end{bmatrix}, \,
    x_k = 
    \begin{bmatrix}
        z \\
        \dot{z}
    \end{bmatrix}$ &
    {\begin{tabular}{l}
        Position (TD, LO)  \\
        Velocity (MS, LO) \\
    \end{tabular}} &
    {\begin{tabular}{l}
        $\gamma_1=0.108$  \\
        $\gamma_2=0.329$ \\
        $\gamma_3=0.594$ \\ 
    \end{tabular}}  \\
\cline{1-1} \cline{3-5}
& KF - Meas. Update & $z_k = H \hat{x}_k + n_a, \, u_k = [0,0,1] \, a_k^{w}$ & \\
KF2 & {\begin{tabular}{l}
    $K = P_k^-  H^T (H  P_k^-  H^T + R)^{-1}$ \\
    $x_k = x_k^- + K  (z_k - H  x_k^-)$ \\
    $P_k = (I - K H)  P_k^-$ \\
\end{tabular}} & 
    $F = \begin{bmatrix}
        1 & \Delta t & -0.5 \Delta t^2 \\
        0 & 1 & -\Delta \\
        0 & 0 & 1
    \end{bmatrix}
    , \,
    G = 
    \begin{bmatrix}
        0.5 \Delta t^2 \\
        \Delta t \\
        0
    \end{bmatrix}$ & 
    {\begin{tabular}{l}
        Position (TD, LO)  \\
        Velocity (MS, LO) \\
        Accel. Bias (Aerial Phases) \\
    \end{tabular}}  &
    {\begin{tabular}{l}
        $\gamma_1=0.111$  \\
        $\gamma_2=0.497$ \\
        $\gamma_3=0.903$ \\ 
    \end{tabular}}\\
    & & $x_k = [z \, \dot{z} \, \ddot{z}_{bias}]^T$ & \\
\hline
& ESKF - Nominal State Prediction & $z_k = H \hat{x}_k + n_a - H x_k^-, \, u_k = [0,0,1] \, a_k^{w}$ & \\
ESKF1 &   {\begin{tabular}{l}
        $x_k^- =  f(x_{k-1}) + g(x_{k-1}) u_k$ \\
        \hspace{-1.8ex}ESKF - Prediction \\
        $P_k^- = F P_{k-1} F^T + Q$\\
    \end{tabular}} &
    \multirow{2}{*}{$F = \begin{bmatrix}
        1 & \Delta t & -0.5 \Delta t^2 \\
        0 & 1 & -\Delta t \\
        0 & 0 & 1
    \end{bmatrix}, \, G = \begin{bmatrix}
        0.5 \Delta t^2 \\
        \Delta t \\
        0
    \end{bmatrix}$} & 
    {\begin{tabular}{l}
        Position (TD, LO)  \\
        Velocity (MS) \\
    \end{tabular}} &
    {\begin{tabular}{l}
        $\gamma_1=0.127$  \\
        $\gamma_2=0.417$ \\
        $\gamma_3=0.724$ \\ 
    \end{tabular}} \\
%
\cline{1-1} \cline{4-5}
& ESKF - Meas. Update & & \\
ESKF2 & {\begin{tabular}{l}
    $K = P_k^-  H^T (H  P_k^-  H^T + R)^{-1}$ \\
    $\delta x_k = \delta x_k^- + K (z_k - H \delta x_k^-)$ \\
    $P_k = (I - K H)  P_k^-$ \\
    $x_k = x_k^- + \delta x_k$ \\
\end{tabular}} & 
{\begin{tabular}{l}
    $x_k = [z \, \dot{z} \, \ddot{z}_{bias}]^T$ \\
    $\delta x_k = [\delta z \, \delta \dot{z} \, \delta \ddot{z}_{bias}]^T$ \\
\end{tabular}} &  
    {\begin{tabular}{l}
        Position (TD, LO)  \\
        Velocity (MS) \\
        Accel. Bias (Aerial Phases) \\
    \end{tabular}} &
    {\begin{tabular}{l}
        $\gamma_1=0.134$  \\
        $\gamma_2=0.423$ \\
        $\gamma_3=0.725$ \\ 
    \end{tabular}} \\[4ex]
\hline
\end{tabular}
\begin{tablenotes}
\item Parameters: state transition matrix, $F$, control matrix, $G$, estimated covariance, $P$, process noise covariance, $Q$, observation matrix, $H$, measurement noise covariance, $R$, Kalman gain, $K$, input, $u_k$, and measurements, $z_k$. Abbreviations and descriptions: Measurement (Meas.), Acceleration (Accel.), Kalman Filter (KF), Error State Kalman Filter (ESKF), true state ($\hat{x}_k$).
\end{tablenotes}
\label{tab:filter}
\end{center}
\end{table*}

\section{State Estimation Methodology}
Two common approaches involve a Kalman filter (KF) with states ($x_k = [p_k,v_k,\theta_k,b_{ak},b_{gk}]$) or an error state Kalman filter (ESKF) with error states ($\delta x_k = [\delta p_k, \delta  v_k, \delta \theta_k, \delta b_{ak}, \delta b_{gk}]$) and nominal states ($x_k$); where, the subscript $k$ is the time step, $p_k$ is the position, $v_k$ is the velocity, $\theta_k$ is attitude, $b_{ak}$ is the accelerometer bias, $b_{gk}$ is the gyroscope bias, and $\delta$ indicates the error between the true and estimated states. The inertial measurement unit (IMU) and accelerometer provide both the linear acceleration and rotational velocity as,
\begin{align}
    a^{IMU} &= \hat{a} + b_a + n_a + \alpha_a \\
    \omega^{IMU} &= \hat{\omega} + b_g + n_g
\end{align}
where, $\hat{a}$ is the true acceleration, $n_a$ is the accelerometer noise, $\alpha_a$ is the accelerometer aliasing, $\hat{\omega}$ is the true rotational rate, and $n_g$ is the gyroscope noise. As with \cite{liu_tlio_2020}, neglecting the Coriolis forces and curvature of the earth, the state transition equations are as follows,
\begin{align}
    &p_{k+1}^w = p_k^w + v_k^w \Delta t + \frac{1}{2} ((R_k^{bw}(a_k^{IMU} - b_{ak}) + g^w) \Delta t^2 \label{eq:st1}\\
    &v_{k+1} = v_k + (R_k^{bw}(a_k^{IMU} - b_{ak}) +g^w) \Delta t \label{eq:st2}\\
    &R_{k+1}^{bw} = R_{k}^{bw} (I + [\omega^{IMU}_k - b_{gk}]_\times \Delta t) \label{eq:st3}\\
    &b_{ak+1} = b_{ak} + n_{ba} \label{eq:st4}\\
    &b_{gk+1} = b_{gk} + n_{bg} \label{eq:st5}
\end{align}
where, the superscript $w$ indicates the world frame, $R^{bw}$ is the rotation matrix from the body to the world frame, $I + [\omega^{IMU}_k - b_{gk}]_\times \Delta t$ is the first-order Taylor expansion of Rodrigues' rotation formula, $a^{IMU}$ is the measured acceleration in the IMU frame, $\omega^{IMU}$ is the measured rotation rate in the IMU frame, the subscript $\times$ indicates the cross-product operator matrix for the bias compensated angular rate measurement, and $g^w$ is the gravity vector.

Developing a KF from the state transition equations \ref{eq:st1}-\ref{eq:st5}, shows that measurement updates impact only the measured state and its time integrals and derivatives; thereby allowing for extraction and independent development of the KF for the vertical states. Extending this to an ESKF, it does show a small amount of coupling between non-integral and non-derivative states, due to the attitude error impacting the velocity error \cite{norrdine_step_2016}. However, assuming small attitude errors and therefore limited coupling, this isolation and development can be applied to both KFs and ESKFs. This allows for a focused study on managing significant and variable aliasing in acceleration measurements (MultiMo-MHR: Nyquist frequency $\sim$420 Hz). The challenge is further compounded by the fact that the most significant aliasing occurs at the regions of maximum energy transfer during TD and LO; as seen in the previously developed model (Fig. \ref{fig:robot}b) and experimental testing \cite{burns_design_2025}.

\subsection{Hopping Vertical State Estimator (HVSE)}
Differing from \cite{Zhu2022, bai_agile_2024}, with thrust-to-weight ratios (TWRs) $\geq$ 1, the MultiMo-MHR operates with a maximum TWR limited to 83.7\%. Operating at TWR $<$ 1 poses an additional challenge as the true \textit{drop} phase transition is determined by the dynamic behavior and not the controller; where an early estimated \textit{drop} phase transition reduces the maximum performance while a late \textit{drop} phase transition reduces hopping efficiency. Vertical state estimation at a TWR of $\sim$90\% has been shown, however, the work required both position and acceleration to estimate vertical velocity \cite{wang_terrestrial_2023,kang_fast_2024}. 


Previous studies in pedestrian tracking employing Kalman Filters (KF) to estimate both position and velocity, have shown that assuming zero velocity of the foot during ground contact can allow for inferred velocity measurements, termed zero-velocity updates (ZUPTs), which can significantly reduce drift in the state estimation \cite{foxlin_pedestrian_2005,wang_stance-phase_2015}. Further studies have added zero-altitude updates \cite{norrdine_step_2016} to reduce the drift in the vertical axis as well. Expanding on the concept for vertical state estimation in high-performance hopping robots, three of the four phase transitions including touchdown (TD), max squat (MS), and liftoff (LO) as well as the motor drive commands allow for inferred-measurements updates (IMUPTs) including: 
\begin{description}
  \item [IMUPT-position (TD,LO)] $= [0,0,1] \, R^{bw}[0,0,-L_f]^T$ at TD and LO where the vertical position is inferred from the known distance between the robot's center-of-mass and the foot ($L_f = (m_B L_3 + m_L L_1)/(m_B+m_L) - (L_1 + L_2)$) without spring extension.
  \item [IMUPT-velocity (MS)] $= 0$ at MS where the vertical velocity is inferred from the body's change in direction.
  \item [IMUPT-velocity (LO)] $= v_{LO} \delta_{vLO}$ at LO where the vertical velocity is inferred from an understanding of the system losses, momentum transfer, and inherent acceleration aliasing at liftoff. The scaling factor $\delta_{vLO} = (c_{vel2} v_{LO}^2 + c_{vel1} v_{LO} + c_{vel0}) (c_{ch1} h_{ch} + c_{ch0})$; where, liftoff velocity estimate, $v_{LO}$, and preset commanded height for the subsequent hop, $h_{ch}$, determine the velocity adjustment (Fig. \ref{fig:state_est}b.1 $\sim$2.06\% mark). The $c_{vel2,vel1,vel0}$, and $c_{ch1,ch0}$ are the $v_{LO}$ and $h_{ch}$ fit coefficients respectively, and account for estimation errors during the stance and subsequent rebound phases.
  \item [IMUPT-acceleration (Aerial Phases)] $= [0,0,1](R^{bw} F_{U1}/m-g^w)$ during the aerial phase where the acceleration is inferred from the total commanded motor thrust, $F_{U1}=c_{m1} (\sum_{i=1}^4 d_{mi}) + c_{m0}$; where $c_{m1,m0}$ are the fit coefficients to convert from duty-cycle to thrust, $d_{mi}$ are the commanded motor duty-cycles, $m$ is the overall robot mass.
\end{description}
where, Appendix Robot Parameters shows the variables.

Based on the identified IMUPTs, two Kalman filters (KF1, KF2) and two error state Kalman filters (ESKF1, ESKF2) are developed (Table \ref{tab:filter}). The vertical acceleration input $u_k = [0,0,1] \, a_k^{w}$ is composed of the world frame acceleration measurement $a_k^{w} = R_k^{bw} (a_k^{IMU} - b_{ak}) + g^w$. These filters required both a process and measure noise covariance matrix containing up to four additional parameters including the standard deviation of the acceleration ($\sigma_{az}$), position ($\sigma_{pz}$), velocity ($\sigma_{vz}$), and bias ($\sigma_{bz}$) measurements. Additional adaptability is included in the state estimation, by first low pass filtering the accelerometer measurements resulting in two cutoff frequency parameters, $f_{HVSE}$ and $f_{HPE}$, for the HVSE and HPE, respectively. Finally, as two accelerometers are used, the switching g-force parameter, $g_{s}$, is included to determine at which point to switch from the low-g to high-g accelerometer measurements. The total Kalman filter parameters are presented in Table \ref{tab:opt_parm}; highlighting the challenge to hand tuning these filters.

\subsection{HPE and HVSE Training}
Initial experimental trials (n=998 hops using the true state for control) were conducted at hopping heights between 1 and 4 meters; where the acceleration readings from both the low-g and high-g accelerometers were captured (robot approx. 840 Hz) along with the true state data from a motion capture system (Vicon Vantage, 600 Hz). Implementing in simulation (MATLAB) a particular HPE and HVSE on the captured experimental data, allows for an exact recreation of what the robot would experience and the associated state estimation, without the risks and time associated with testing on the real robot. Therefore, assuming the initial experiments span the complete range of desired robot operational characteristics, this allows for a complete understanding of the HVSE error including all environmental (e.g. wind, substrate properties), locomotion (e.g., hopping heights, traveling speeds), control (e.g., controller driven behaviors), and sensor (e.g., noise, positioning) characteristics. 

The ability to capture large amounts of experimental data and exactly simulate the HVSE errors, suggests that the Kalman filter parameters can be learned from the data. Therefore, due to the challenging parameter space with significant local minima, a genetic algorithm was developed to optimize the HPE and HVSE parameters listed in Table \ref{tab:opt_parm}. To control the hopping height of the robot, the most important parameter is the hop apex height, $z_{HA}$. Therefore, the cost function ($\mathcal{L}_{c}$) is set as the mean-absolute-percentage-error (MAPE, $\gamma_1$) in the apex height as,
\begin{align}
    &\text{min} \, \mathcal{L}_{c} =
    \begin{cases}
        w_1 \gamma_1 & \text{if  } n_{HA}= \hat{n}_{HA} \\
        w_2 \gamma_2 + w_3 \gamma_3 & \text{otherwise}
    \end{cases} \\
    &\gamma_1 =  \frac{1}{n_{HA}} \sum_{i}^{n_{HA}}  \bigg| \frac{z_{HA_i} - \hat{z}_{HA_i}}{\hat{z}_{HA_i}} \bigg| \label{eq:g1}\\
    &\gamma_2 =  \sqrt{\frac{1}{n_p} \sum_{i}^{n_p}   (z_i - \hat{z}_i)^2 }, \,
    \gamma_3 =  \sqrt{\frac{1}{n_v} \sum_{i}^{n_v}   (\dot{z}_i - \hat{\dot{z}}_i)^2 }  \label{eq:g3}
\end{align}
where, the ground true values [$\hat{z}_{HA},\hat{z},\hat{\dot{z}}$] of the hop apex height, $z_{HA}$, vertical position, $z$, and vertical velocity, $\dot{z}$, are as shown respectively. The RMSE in both vertical position, $\gamma_2$, and velocity, $\gamma_3$, are used to maintain a gradient in the event that the estimation does not capture the correct number of hops. The number of estimated hop apexes, $n_{HA}$, the true hop apexes, $\hat{n}_{HA}$, vertical heights, $n_p$, and vertical velocities, $n_v$, are represented accordingly. Finally, the weights are as follows, $[w_1=100, \, w_2=w_3=10]$.

The genetic algorithm is initialized as follows: individuals = 15 parameters (Table \ref{tab:opt_parm}), initial population = 1000 randomly generated individuals, maximum generations $n_{ga}$ = 20, elites carried over to the subsequent generation = 5\%, crossover fraction = 80\%, and the mutation fraction = 15\%. Stochastic universal sampling is used to select two parents and the children are created through a uniform scattering crossover; where each optimization parameter is given a random one (parent-1) or zero (parent-2), determining which parent provides the parameter for the child. Finally, a constraint-aware adaptive mutation algorithm is used that creates the child as follows, $x_{ga}^\prime = x_{ga}+\alpha_0 \, \exp(n_{ga_i}/n_{ga}) \, d/||d||$; where the direction vector, $d_{(15\times1)}$, is a random vector in which each component is drawn from a normal distribution $\sim \mathcal{N}(0,1)$. The $n_{ga_i}$ is the current generation number, and $\alpha_0$ is the initial step size that can be reduced if the constrains are violated. Finally, as stated previously, learning the parameters requires that the training data covers all desired robot operational characteristics. However, initial testing has shown that a much more limited set of data than that collected sufficiently well cover the operating range allowing for more efficient optimization on a reduced set containing 111 hops spread across the [1,2,3,4] m hop trials. This results in a reduction from multiple days per optimization cycle, to approximately 5 hours of parallel processing across eight cores; where code optimization could improve this further.

\begin{table}[tbp!]
\caption{Optimization Parameters}
\begin{center}
\begin{tabular}{p{0.05\textwidth}p{0.16\textwidth}p{0.05\textwidth}p{0.07\textwidth}p{0.045\textwidth}}
\hline
\textbf{Parm.} & \textbf{Description} & \textbf{Opt. Value} & \textbf{Bounds} & \textbf{Units} \\[-1ex]
& &  &  \\
\hline
$f_{HVSE}$ &  Cutoff Freq. (HVSE)& 7 &[400,5] & Hz \\
$f_{HPE}$ & Cutoff Freq. (HPE) & 8 &[400,5] & Hz \\
$g_{s}$ & Accel. Switching g-force & 14.24 &[14.5,12] & g-force \\
$\sigma_{az}$ & Accel. Meas. STD  & 9.9857 &[10,0.0001] & m/s$^2$ \\
$\sigma_{bz}$ & Accel. Bias Meas. STD & NA &[10,0.0001] & m/s$^2$ \\
$\sigma_{vz}$ & Velocity Meas. STD & 9.5722 &[10,0.0001] & m/s \\
$\sigma_{pz}$ & Position Meas. STD & 0.0091 &[10,0.0001] & m \\
$c_{vel2}$ & Fit Constant Coef. vel$^2$ & -1.1246 &[10,-10] &  \\
$c_{vel1}$ & Fit Constant Coef. vel$^1$ & 5.9203 &[10,-10] &  \\
$c_{vel0}$ & Fit Constant Coef. vel$^0$ & 6.7054 &[10,-10] &  \\
$c_{ch1}$ & Fit Constant Coef. ch$^1$ & 6.2138 &[10,-10] &  \\ 
$c_{ch0}$ & Fit Constant Coef. ch$^0$ & 8.4355 &[10,-10] &  \\ 
$c_{m0}$ & Fit Constant Coef. $d_{mi}^1$ & NA &[10,-10] &  \\
$c_{m1}$ & Fit Constant Coef. $d_{mi}^0$ & NA &[10,-10] &  \\
\hline
\end{tabular}
\begin{tablenotes}
\item Abbreviations: Measurement (Meas.), Acceleration (Accel.), standard deviation (STD), coefficient (Coef.).
\end{tablenotes}
\label{tab:opt_parm}
\end{center}
\end{table}

\subsection{Training Results}
Optimizing across the four filters (KF1, KF2, ESKF1, ESKF2) for the MAPE in the apex height ($\gamma_1$) results in KF1 with the lowest error parameters (Table \ref{tab:filter} Error column). The addition of the accelerometer bias measurement (IMUPT-Accel. Bias (Aerial Phase)) to KF2 results in increased errors. This suggests that the measurement adds more uncertainty than useful information; likely due to the delay between motor command and thrust production. Shifting to ESKF1 and ESKF2, IMUPT-Velocity(LO) was removed because of an observed instability suggesting that the short-term significant change in the error, cause by the aliasing at TD and LO, is not well managed by the ESKF.

Selecting KF1, the optimized parameter values can be seen in Table \ref{tab:opt_parm}, which result in a MAPE in the apex height of 10.77\% (n=998 hops) over the complete set of training data. Analyzing the optimized values in more detail, we can further understand the system's characteristics. First, the cutoff frequencies for both the HVSE and HPE show a trend towards low values, causing significant phase lag and smoothing of the acceleration measurements. This suggests a preference towards removing the inherently arbitrary aliasing of the acceleration measurements over phase accuracy for both the HVSE and HPE; where the colored dots in Fig. \ref{fig:position_velocity_error} show the HPE's transitions during the subsequent experimental trials. Second, the liftoff velocity and commanded height show a correlation with the HVSE error as the exponential coefficients ($c_{vel2}$, $c_{vel1}$, $c_{ch1}$) are not equal to zero. Third, as expected, the smoothing does show an increase in the average error as evident by the non-unity constant coefficient values ($c_{vel0}$, $c_{ch0}$). Finally, arbitrary initialization of the estimated covariance matrix, $P$, yielded poor initial state estimations at the first TD. Therefore, to improve the initial state estimation and accelerate convergence of the KF, the $P$ is initialized to an average $P$ after convergence; yielding a much closer estimation of the real value for the first TD (MultiMo-MHR: initial $P = [0.0582, 0.0774; 0.0774, 0.1441] .* 10^{-4}$).

\begin{figure}[tbp]
\centerline{\includegraphics[width=0.5\textwidth]{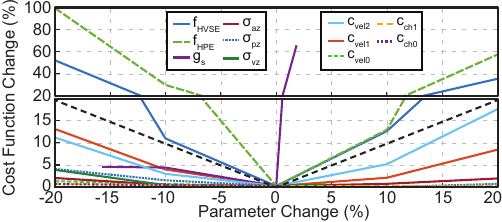}}
\caption{Sensitivity to changes in the optimized parameters; $\mathcal{L}_{c} = 100 \gamma_1 + 10 \gamma_2 + 10 \gamma_3$ from equations \ref{eq:g1}, \ref{eq:g3}.}
\label{fig:sensitivity}
\end{figure}

\begin{figure}[tbp]
\centerline{\includegraphics[width=0.5\textwidth]{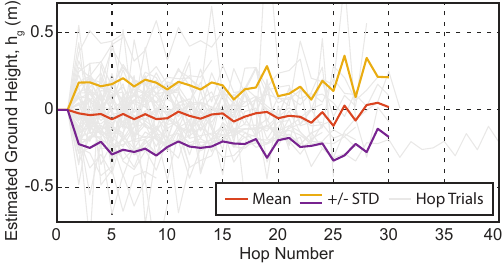}}
\caption{Change in the estimated ground height over multiple hops; where STD is the standard deviation, and the true ground height is zero.}
\label{fig:ground}
\end{figure}

Figure \ref{fig:sensitivity} shows the sensitivity of the optimized parameters to variation, where the cutoff frequencies ($f_{HVSE}$, $f_{HPE}$) and the switching g-force parameter ($g_s$) are the most sensitive; resulting in larger percent changes to the cost function than the percent change to the parameter. Furthermore, the significant change caused by increasing the $g_s$, suggests that the low-g accelerometer's noise characteristics change abruptly as its limits are approached, and that the optimization methodology can learn the sensor characteristic from the data.

As can be seen in \cite{norrdine_step_2016}, zeroing altitude (IMUPT-position) removes drift but also the ability to track absolute height in environments where the ground is not flat; resulting in an error relative to the previous contact surface of approximately 0.05 m per step. Absolute height can be recovered through summing the change in height of each stepping or hopping cycle prior to the zeroing. The change in the ground height between cycles can be calculated in two ways including,
\begin{align}
    \Delta h_{1n} &= (h_{HAn}-h_{TDn}) - (h_{HAn-1}-h_{TDn-1}) \\
    \Delta h_{2n} &= h_{TDn}
\end{align}
where the estimated height at the hop apex ($h_{HA}$) and TD ($h_{TD}$) are represented accordingly, and $n$ indicates the hop number. The difference between them ($h_{HA}-h_{TD}$) gives a measure of the drop height, and the difference between subsequent drop heights, provides a measure of the change in the ground height ($\Delta h_1$). Second, the estimated TD height ($h_{TD}$) provides a direct measure of the change in the ground height ($\Delta h_2$). Taking the average of these two measures, 
\begin{align}
    h_{gn} &= h_{gn-1} + (\Delta h_{1n} + \Delta h_{2n})/2
\end{align}
yields an estimate of the change in the ground height for the subsequent hop or step ($h_g$). Comparing the estimated ground height error of KF1 (Figure \ref{fig:ground}) to that of an untrained KF \cite{norrdine_step_2016}, shows that not only can drift be significantly reduced but that the absolute ground height can be accurately tracked. With an overall average ground height error of 0.018 m after 30 hops of up to 4 meters, the trained KF1 can achieve a 64\% improvement on the single step error and significantly better than the sum over the same 30 steps. This is due to the optimization tending towards zero mean error with minimum standard deviation; resulting in a minimum error with an equal chance to be positive or negative for each hop.

\begin{figure}[tbp]
\centerline{\includegraphics[width=0.5\textwidth]{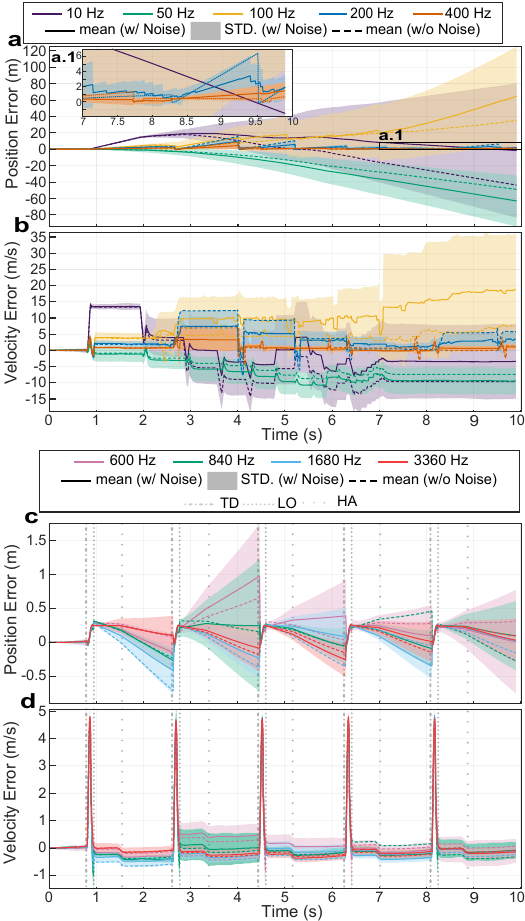}}
\caption{Simulated State Estimation Error. a) Unbound and large bound position error, given state estimation at 10 to 400 Hz. b) Unbound and large bound velocity error, given state estimation at 10 to 400 Hz. c) Bound position error, given state estimation at 600 to 3360 Hz. b) Bound velocity error, given state estimation at 600 to 3360 Hz.}
\label{fig:sim}
\end{figure}

\section{Sensing Frequency}
The previously developed dynamic model in \cite{burns_design_2025} can be used to understand the impact of the sensing frequency on the HVSE; which fundamentally controls the aliasing of the accelerometer measurements. Specifically, the state estimation error boundedness and growth as a function of the limiting frequency; that being sensor read speeds or state estimation frequency. To explore the estimation error behavior, the measurement and estimation frequencies are varied over 12 values including [3360, 1680, 840, 700, 600, 500, 400, 300, 200, 100, 50, 10] Hz; where the MultiMo-MHR's HVSE is optimized to, and operates at, 840 Hz. The simulation (MATLAB) results are shown in Fig. \ref{fig:sim} with a control frequency of 400Hz, commanded height, $h_{ch}$, of 3 m, and the robot parameters in Appendix Robot Parameters. The results are shown for simulations both with and without sensor noise, and the controller is using the true state data, at the measurement frequency, as reference for control. The simulation without sensor noise (mean (w/o Noise)) can highlight effects of aliasing alone while the simulation with noise (mean and STD. (w/ Noise)) can provide an understanding of the true error growth characteristics.

As can be seen in Fig. \ref{fig:sim}a,b, frequencies from 10 to 100 Hz show an unbounded state estimation error with significant growth rate in the standard deviation as indicated by the large monotonic increase over time (Fig. \ref{fig:sim}a,b STD. (w/ Noise)); at $\sim$7 seconds both the 10 and 50 Hz trials are unable to sustain hopping and simply remain stationary on the ground. This unboundedness is similar to the drift associated with direct accelerometer integration (DAI) and therefore, provides no improvement in state estimation over DAI. 

Frequencies from 200 to 400 Hz show a bound error characteristic in which the standard deviation in the error is not increasing monotonically (Fig. \ref{fig:sim}a,b STD. (w/ Noise)). However, both mean and standard deviation remain sufficiently large so as to cause variation in the subsequent hops (Fig. \ref{fig:sim}a,b mean and STD. (w/ Noise)). 

Frequencies from 600 to 3360 Hz show both a bound mean error and a significant reduction in the standard deviation of the error, in both position and velocity, within each hop cycle (Fig. \ref{fig:sim}c,d mean and STD. (w/ Noise)); creating a deterministic starting point for the subsequent hop. Additionally, from Fig. \ref{fig:sim}c,d, we see both a low sensitivity to minor changes in frequency and diminishing returns as the frequency is increased. This suggests that the impact of accelerometer aliasing is saturating, and that the HVSE, trained at 840 Hz, is robust to frequency changes in the range of 600 to 3360 Hz. Interestingly, as seen in Fig. \ref{fig:sim}d, it is not required that the mean error in velocity be reduced; only that the standard deviation be reduced as the velocity scaling factor ($\delta_{vLO}$) accounts for mean errors in velocity.

\section{Hopping Height Control Experiments}
Hopping experiments were conducted with the MultiMo-MHR (Fig. \ref{fig:robot}a) at target heights of 1, 2, 3, and 4 m. Trials had two possible measurement inputs for vertical control; true state (GT, motion capture system, n = 70 hops) or the HVSE position and velocity (n = 121 hops). The HVSE is recorded in all scenarios to identify if the HVSE and vertical controller are decoupled and to evaluate the estimator performance. 

\begin{figure*}[tbp]
\centerline{\includegraphics[width=0.9\textwidth]{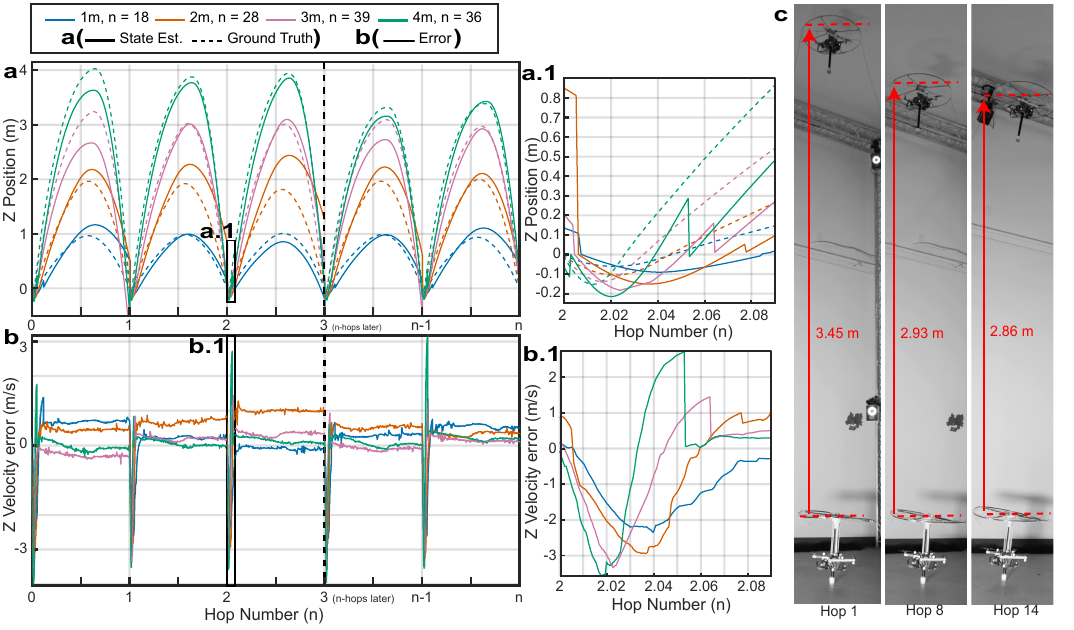}}
\caption{Example hopping trials using the hopping vertical state estimator (HVSE) based control a-b) Displays the vertical position and velocity error of the first 3 and last 2 hops for hopping trials at heights of 1,2,3\&4 m. Each has a zoomed in region (a.1, b.1) showing the stance phase behavior. c) Snapshots from a high-speed video of the MultiMo-MHR hopping with the HVSE at a desired height of 3.25 m (Supplemental Video: Video1).}
\label{fig:state_est}
\end{figure*}

The experimental procedure begins with zeroing of the accelerometers and motion capture system with the robot positioned vertically on the ground. The robot is then flown to the set initial height with a separate controller (only used in the startup procedure) before entering the first drop state, beginning the hop cycle and activating the hopping height controller. For a complete breakdown of the hop cycle and corresponding HPE and HVSE see Fig. \ref{fig:stateMachine}; where, Fig. \ref{fig:robot} and \ref{fig:state_est} provide a visual reference of the hop cycle and HVSE.

\subsection{Results}
\begin{figure*}[tbp]
\centerline{\includegraphics[width=1\textwidth]{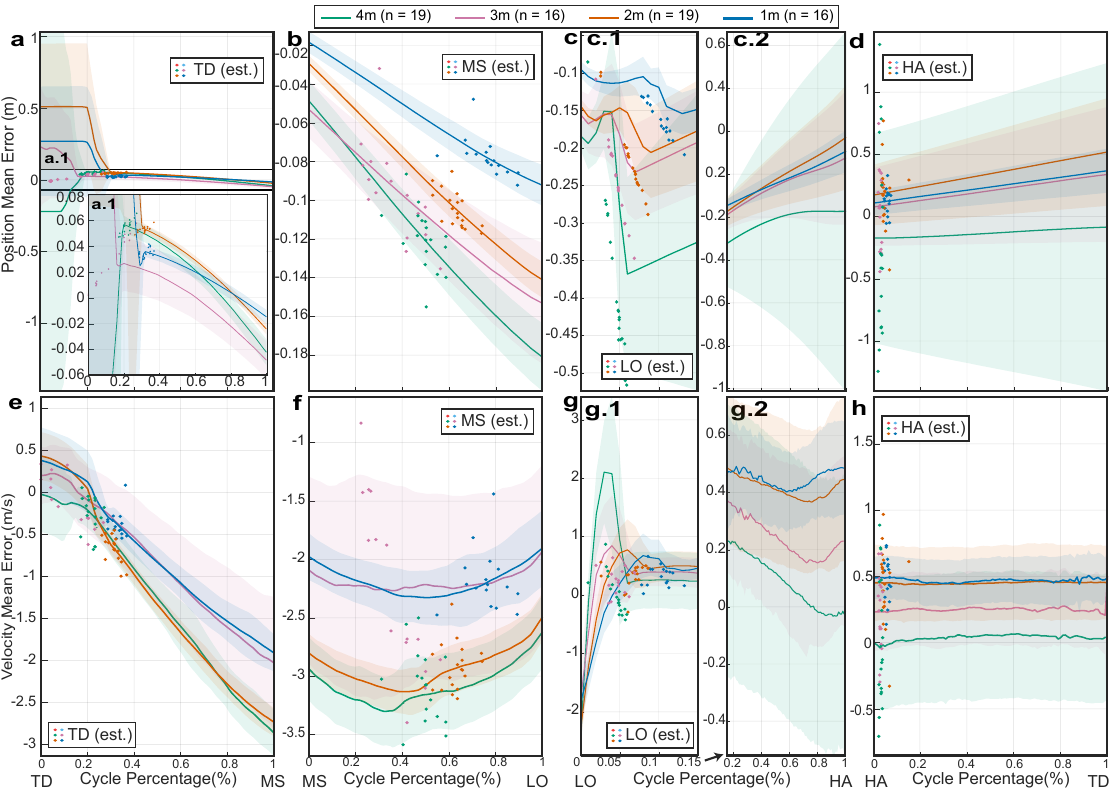}}
\caption{Mean and standard deviation of the position (a-d) and velocity (e-h) error  for the experimental trials (n=70 hops) utilizing the true state (motion capture data) for height control. a\&e) \textit{Stance down} phase, TD to MS. a\&e.1) Zoomed in region of (a\&e) to show the behavior after TD). b\&f) \textit{Stance up} phase, MS to LO. c\&g) \textit{Rebound} phase, LO to HA. c\&g.1) Zoomed in region from 0-15\% of \textit{Rebound} phase c\&g.2) Zoomed in region from 15-100\% \textit{Rebound} phase. d\&h) \textit{Drop} phase, HA to next TD. Note) The phases are segmented by the true state data, while the dots indicate the points at which the robot identified the specified transition.}
\label{fig:position_velocity_error}
\end{figure*}

Figure \ref{fig:position_velocity_error} shows the experimental results of the vertical position and velocity estimation using the HVSE. Table \ref{tab:error_tab_v2} compares the HVSE to other vertical position and velocity estimation techniques used in hopping robots across five error metrics (Appendix: Error Metrics) including: the mean of the normalized MAE in the vertical position (M1), the mean of the normalized MAE in the vertical velocity (M2), mean absolute percent error (MAPE) in the apex height (M3), mean absolute error (MAE) in the apex time (M4), and MAE in the desired apex height (M5). The table compares three vertical state estimation techniques to the HVSE. These include a Newtonian approach that assumes a ballistic aerial trajectory where the LO state is determined by a leg encoder (BA1) \cite{Raibert1984,Yim2020,bai_agile_2024}. However, as rotor based hopping robots may not follow a ballistic trajectory, a natural extension is to dead-reckoning the accelerometer measurements (LO to TD) to improve the overall estimates (DR1). Finally, as shown in \cite{foxlin_pedestrian_2005,wang_stance-phase_2015,norrdine_step_2016}, a Kalman filter with a zero altitude update at TD can be employed to further improve the state estimation (KF3 = KF1 with only IMUPT-position(TD)). Four variations of the HVSE are shown including HVSE1, using the true state (GT) for control, and HVSE2, using the estimated state (SE) for control; where HVSE3 and HVSE4 only include the aerial phase errors. The controller for each technique can use either the true state (GT), focusing on the estimation characteristic, or the estimated state (SE), focusing on the overall hop height control performance. 

Validating the simulation (Sim.) data, the experimental (Exp.) results show a similar reduction in both the error mean and standard deviation in position estimation at TD; as a result of the IMUPT-Position(TD) ((Sim.) Fig. \ref{fig:sim}c, (Exp.) Fig. \ref{fig:position_velocity_error}a, a.1). The IMUPT-Velocity(MS) in both cases reduces the velocity mean error growth rate but with limited impact on the standard deviation ((Sim.) Fig. \ref{fig:sim}d, (Exp.) Fig. \ref{fig:position_velocity_error}f). Finally, in both cases at LO (IMUPT-Position(LO), Velocity(LO)), a significant reduction in the velocity error's mean and standard deviation is observed ((Sim.) Fig. \ref{fig:sim}d, (Exp.) Fig. \ref{fig:position_velocity_error}g.1). The significant reductions in both the mean and standard deviations in the position and velocity errors creates a deterministic starting point for the subsequent hop cycle estimation; minimizing the error cause by the significant aliasing of the accelerometer measurements. 

Table \ref{tab:error_tab_v2} highlights a number of key concepts. First, comparing the common hopping robot vertical state estimation technique (BA1) to its natural extension for rotor based hopping (DR1), suggests that the ballistic trajectory assumption may not be valid. Second, comparing DR1 to the Kalman filter based pedestrian navigation technique (KF3), suggests that a leg encoder can achieve similar results to that of a zeroing altitude condition at TD in a Kalman filter based approach. Third, comparing KF3 to the HVSE, suggests that a trained Kalman filter with the specified IMUPTs can improve the state estimation under significant aliasing conditions. Fourth, comparing HVSE1 to HVSE2 only shows a significant different in M5, suggesting that the controller and HVSE are independent; however, as expected M5 does show a significant difference, suggesting that the HVSE does introduce error beyond that caused by the controller itself. Fifth, comparing HVSE1 and HVSE2 to HVSE3 and HVSE4, suggests that the estimation of the stance phase states is more challenging; however, as the vast majority of the motion of the robot occurs in the aerial phase this will have a minimal overall impact. Sixth, from basic optics, the M1 metric is shown to be good approximation of the error in the horizontal state estimation, given that an optical flow sensor's range finder (height measurement) is replaced with the HVSE (Appendix: Optical Flow Error). Seventh, given the MultiMo-MHR travels up to $\sim8$ meters at velocities that range between $\sim\pm7$ m/s in $\sim$2.6 seconds coupled with the significant and stochastic nature of aliasing, a certain amount of error is inherent without further direct measurements of the states. Finally, using experimental data, a KF or ESKF can be trained to extract as much information as possible from a highly aliased signal, and be incorporated back into a full state KF or ESKF.

\begin{table}[tbp!]
\caption{HVSE Performance Metrics}
\begin{center}
\begin{tabular}{p{0.08\textwidth}p{0.04\textwidth}p{0.035\textwidth}p{0.035\textwidth}p{0.025\textwidth}p{0.135\textwidth}}
\hline
\textbf{Metric} & \textbf{Est.} & \textbf{Meas.} & \textbf{1-4m} & \textbf{Units} & \textbf{Statistics \footnote[3]{}} \\[-1ex]
& &  &  \\
\hline
(M1) Mean & BA1& GT& 196.8& \%    &t(138)=30.97, p=0.00 \\
nMAE Traj& DR1& GT& 44.05& \%    &t(138)=5.62, p=0.00 \\ 
Pos. & KF3& GT& 60.45& \% & t(138)=7.49, p=0.00\\ 
& HVSE1& GT& 21.24 & \% & t(189)=0.99, p=0.32   \\
& HVSE2& SE& 19.46 &\% &Comparison \\
& HVSE3\textsuperscript{\dag}& GT& 20.21& \% &\\
& HVSE4\textsuperscript{\dag}& SE& \cellcolor[HTML]{C0C0C0}\textbf{18.97}& \% &\\ 
\hline
(M2) Mean & BA1& GT& 179.9& \%    &t(138)=24.15, p=0.00 \\
nMAE Traj& DR1& GT& 28.15& \%    &t(138)=3.17, p=0.00 \\ 
Velo. & KF3& GT& 37.84& \% & t(138)=5.24, p=0.00\\ 
& HVSE1& GT& 19.58 & \% & t(189)=0.31, p=0.76   \\
& HVSE2& SE& 19.11 &\% &Comparison \\
& HVSE3\textsuperscript{\dag}& GT& 17.08& \% &\\
& HVSE4\textsuperscript{\dag}& SE& \cellcolor[HTML]{C0C0C0}\textbf{16.50} & \% &\\ 
\hline
(M3)& BA1& GT&  53.85& \%    &t(138)=32.01, p=0.00 \\
MAPE& DR1& GT&  43.36& \%    &t(138)=5.90, p=0.00 \\ 
Apex& KF3& GT&  50.11& \% & t(138)=7.38, p=0.00\\ 
Height& HVSE1& GT&  13.77& \%    &t(189)=0.83, p=0.41 \\
& HVSE2& SE&  \cellcolor[HTML]{C0C0C0}\textbf{12.49}& \%    & Comparison \\ 
\hline
(M4)& BA1& GT&  0.7140& s    & t(138)=16.83, p=0.00 \\ 
MAE& DR1& GT&  0.1323& s    &t(138)=2.66, p=0.00 \\ 
Apex Time& KF3& GT&  0.1388& s & t(138)=4.18, p=0.00\\ 
& HVSE1& GT& 0.0641& s    & t(189)=0.96, p=0.34\\
& HVSE2& SE&  \cellcolor[HTML]{C0C0C0}\textbf{0.0588}& s    & Comparison \\ 
\hline
(M5) MAE& HVSE1& GT& 0.0912& m    &t(170)=5.87, p=0.00 \\
Apex Height&  HVSE2& SE& \cellcolor[HTML]{C0C0C0}\textbf{0.2597}& m    & Comparison \\ 
(desired)&&&&& \\
\hline
\end{tabular}
\begin{tablenotes}
\item Error metrics M1-M5 (n=191 hops, Appendix Error Metrics); where \textit{Meas.} column indicates the measurement used for control (true state (GT), estimated state (SE)). \textsuperscript{*} Values indicate the mean of the RMSE value for a complete hop cycle; TD to subsequent TD. \textsuperscript{\dag} Data uses only the aerial phase as detected by the robot for analysis. \footnote[3]{} Comparing the 1-4 m data to the HVSE2 using a two-tailed t-test given the null hypothesis with a 95\% confidence interval. Related works, BA1 \cite{Raibert1984,Yim2020,bai_agile_2024}, KF3 \cite{foxlin_pedestrian_2005,wang_stance-phase_2015,norrdine_step_2016}.
\end{tablenotes}
\label{tab:error_tab_v2}
\end{center}
\end{table}

\subsection{Agility Metric}
The commonly used hopping performance metric, vertical jumping agility ($\nu_{VJA}$), defined as an average climb speed in a gravitational environment, allows for quantifying and comparing jumping and hopping robot's approximate average locomotion speed \cite{Haldane2016}. The metric is defined as follows, $\nu_{VJA} = \frac{h_{1}}{t_{apogee}} = \frac{h_{1}}{t_{s} + t_{r}}$ where, the apex height, $h_1$, stance through apex time, $t_{apogee}$, stance time, $t_s$, and \textit{rebound} time, $t_r$, define the metric. However, being focused on climb rate, the $\nu_{VJA}$ is unable to account for the \textit{drop} phase time of continuous hopping robots. A subsequent group amended this metric to the hopping agility metric ($\nu_{HA}$) adding the \textit{drop} phase time, $t_d$, to quantify the overall average locomotion speed of continuous hoppers. The metric is defined as follows \cite{bai_agile_2024}, $\nu_{HA} = \frac{2h_{1}}{t_{cycle}} = \frac{2h_{1}}{t_{s} + t_{r} + t_{d}}$ where, the total cycle time is $t_{cycle} = t_{s} + t_{r} + t_{d}$. However, the base metric, $\nu_{VJA}$, was developed to quantify existing systems' performance from videos, and not as a robot design tool. Therefore, while increasing height and decreasing cycle time can increase both metrics, little can be said about control schemes, system losses, and non-constant hopping heights. 

The similarity between the metrics allows for their unification, and a mathematical re-derivation allows for a deeper understanding of the effect of system design and operation parameters (Appendix: Hopping Agility). The unified hopping agility (UHA) metric is as follows: 
\begin{align}
    \nu_{UHA} &= \frac{h_{1} + \beta{h_{0}}}{t_{s} + t_{r} + \beta t_{d}} \label{eq:nu_base}\\
    &= \frac{h_{1} + \beta{h_{0}}}{t_{s} + \sqrt{\frac{2h_{1}}{g(1 - \gamma_{r} + \gamma_{lr})}} + \beta\sqrt{\frac{2h_{0}}{g(1 - \gamma_{d} - \gamma_{ld})}} }\label{eq:nu_expand} \\
    h_{0} &= \zeta_{s}^2\frac{h_{1}(1 - \gamma_{r} + \gamma_{lr})}{(1 - \gamma_{d} - \gamma_{ld})} \label{eq:nu_h0} 
\end{align}
where, the previous hop height, $h_0$, and gravitational acceleration, $g$, are represented accordingly. The average \textit{rebound} phase acceleration, $ a_{rebound}=-g(1 - \gamma_{r} + \gamma_{lr})$, and the average \textit{drop} phase acceleration, $a_{drop}=-g(1 - \gamma_{d} - \gamma_{ld})$, are composed of both energy input, $[\gamma_r, \gamma_d]<1$, representing average thrust forces in \textit{rebound} and \textit{drop}, and energy loss, $[\gamma_{lr}, \gamma_{ld}] \geq 0$, representing average drag forces in \textit{rebound} and \textit{drop}; where positive $[\gamma_r, \gamma_d]$ indicates upward thrust and $[\gamma_{lr}, \gamma_{ld}]$ are always opposite the velocity vector. The change in stance phase velocity between TD and LO, $\zeta_s>0$, can capture energy lost, $0<\zeta_s<1$, no energy change, $\zeta_s=1$, and energy gained, $\zeta_s>1$ in the stance phase. The previous vertical jumping agility metric ($\nu_{VJA}$) can be derived from $\beta=0$ whereas, the hopping agility metric ($\nu_{HA}$) can be derived from $\beta=1$, thus unifying the two metrics within a mathematical framework to inform future hopping robot design. Comparing the MultiMo-MHR to existing hopping robots (Table \ref{tab:comp_agi}) shows that it is able to outperform not only all existing robots but the Galago animal as well.

\begin{table}[tbp!]
\caption{Hopping Platforms Vertical Agility}
\resizebox{\columnwidth}{!}{%
\begin{tabular}{lllllll}
\hline
\multicolumn{1}{|l|}{\textbf{Platform}} & \multicolumn{1}{l|}{\textbf{Hop}} &\multicolumn{1}{l|}{\textbf{$t_{apogee}$}}& \multicolumn{1}{l|}{\textbf{$t_{cycle}$}}& \multicolumn{1}{l|}{\textbf{$\nu_{VJA}$}} & \multicolumn{1}{l|}{\textbf{$\nu_{HA}$}} \\ 
\multicolumn{1}{|l|}{\textbf{}} & \multicolumn{1}{l|}{\textbf{Height(m)}}& \multicolumn{1}{l|}{\textbf{(s)}}& \multicolumn{1}{l|}{\textbf{(s)}}& \multicolumn{1}{l|}{\textbf{(m/s) \cite{Haldane2016}}} & \multicolumn{1}{l|}{\textbf{(m/s) \cite{bai_agile_2024}}} \\ \hline
Salto\cite{Haldane2016}              & 1.008& 0.58\footnote[1]{}& 1.03\footnote[1]{}& 1.75& 1.96\\       
Salto-1P\cite{haldane_repetitive_2017}& 1.252& 0.68\footnote[1]{}& 1.15\footnote[1]{}& 1.83& 2.18\\   
Galago\cite{Haldane2016}& 1.742& 0.78\footnote[1]{}& 1.55\footnote[1]{}& 2.24& 2.24\\   
PogoDrone\cite{Zhu2022}                             & 0.7& 4.18\footnote[2]{}& 8.25\footnote[2]{}& 0.17& 0.17\\   
PogoX\cite{wang_terrestrial_2023}                   & 0.6& 0.44\footnote[2]{}& 0.88\footnote[2]{}& 1.37& 1.37\\   
HopCopter\cite{bai_agile_2024}        & 1.63& 0.71\footnote[3]{}& 1.37\footnote[3]{}& 2.30& 2.37\\ 
\textbf{MultiMo-MHR\footnote[4]{}} & \textbf{3.92}& 1.58& 2.60& \textbf{2.47}& \textbf{3.01}\\ 
\textbf{MultiMo-MHR\footnote[5]{}} & \textbf{4.02}& 1.85& 2.86& \textbf{2.18}& \textbf{2.81}\\  \hline
\\ \vspace{-12pt}
\end{tabular}%
}
\begin{tablenotes}
\item Note: Vertical Jumping Agility ($\nu_{VJA}$) \cite{Haldane2016}. Hopping Agility ($\nu_{HA}$) \cite{bai_agile_2024}. \footnote[1]{}: assumes drop time equals rise time given ballistic motion,\footnote[2]{}: estimated by paper data,\footnote[3]{}: assumes rise and drop time are half the total air time,\footnote[4]{}: MultiMo-MHR max agility,\footnote[5]{}: MultiMo-MHR max achieved height.
\end{tablenotes}
\label{tab:comp_agi}
\end{table}

\section{Summary}
Here we have shown a light-weight, low computational load, IMU only, Kalman filter-based hopping vertical state estimator (position, velocity) coupled to a hopping phase estimator (drop, stance down, stance up, rebound) and the training procedure to learn the specific characteristics of the target system, including significant aliasing of the acceleration measurements. As discussed in the paper, the resulting filter can be reincorporated into a full state Kalman filter and error state Kalman filter. Finally, unlike previous works that trade absolute for relative height estimation to remove drift, the characteristics of the learned filter result in both an absolute height estimation and a significantly reduced drift (average ground height drift of 0.018 m after 30 hops of up to 4 m). The filter was experimentally validated on the MultiMo-MHR with hops up to 4.02 m, 2.47x greater than the next highest continuous hopping robot, at velocities approaching $\pm 7$ m/s. This results in a mean absolute percent error in hop apex height (M3) of 12.5\% with an aerial trajectory average normalized mean absolute error in position (M1) and velocity (M2) of 19\% and 16.5\%, respectively.


\renewcommand{\thesection}{S\arabic{section}}
\renewcommand{\thetable}{S\arabic{table}}
\renewcommand{\thefigure}{S\arabic{figure}}
\renewcommand{\theequation}{S\arabic{equation}}
\setcounter{section}{0}
\setcounter{table}{0}
\setcounter{figure}{0}
\setcounter{equation}{0}

\section*{Appendix}

{\appendices
\section*{Hopping Agility}
Expanding upon the Vertical Jumping Agility ($\nu_{VJA}$)\cite{Haldane2016} and Hopping Agility ($\nu_{HA}$)\cite{bai_agile_2024} metrics, we developed a unified agility metric (UHA: Unified Hopping Agility). Using plots from \cite{Haldane2016}, the hopping cycle for the agility metric proceeds from the prior HA to current HA. While this is irrelevant for constant hopping height cycles, it can create a critical difference in agility for non-constant hopping height cycles. 

Using fundamental Newtonian equations and starting with the sequence of prior HA to TD, the fundamental equations are,
\begin{align}
    a_{drop} &= (-g+a_{d}+a_{ld}) = -g(1 - \gamma_{d} - \gamma_{ld})\label{eq:a_drop}\\
    v_{TD} &= v_{0} + a_{drop} t_{d}\label{eq:vh0} \\
    h_{TD} &= h_{0} + v_{TD}t_{d}+ \frac{1}{2}a_{drop}t_{d}^2 \label{eq:h0}
\end{align}
where we have a control input in the \textit{drop} phase of $a_{d} = g\gamma_{d}$ and accompanying loss shown as $a_{ld} = g\gamma_{ld}$. The system starts at an initial height and velocity ($h_{0}, v_{h0}$) proceeds to a touchdown position and velocity of ($h_{TD}, v_{TD}$) over a \textit{drop} phase time window ($t_{d}$). In the stance phase, between TD and the subsequent LO, the total energy may change do to loss or addition,
\begin{align}
    \zeta_{s} = \frac{v_{LO}}{-v_{TD}} \label{eq:zeta_s}
\end{align}
over a time period ($t_s$). The last segment is the rebound phase, between LO and the current HA, in which the equations are as follows,
\begin{align}
    a_{rebound} &= (-g+a_{r}-a_{lr}) = -g(1 - \gamma_{r} + \gamma_{lr})\label{eq:a_rise}\\
    v_{1} &= v_{LO} + a_{rebound} t_{r}\label{eq:vh1} \\
    h_{1} &= h_{LO} + v_{LO} t_{r}+ \frac{1}{2}a_{rebound} t_{r}^2 \label{eq:h1}
\end{align}
where we have a control input in the \textit{rebound} phase of $a_{r} = g \gamma_{r}$ with accompanying loss shown as $a_{lr} = g\gamma_{lr}$. The system starts at LO with an initial height and velocity ($h_{LO}, v_{LO}$) and continues to an ultimate height and velocity of ($h_{1}, v_{1}$) over a \textit{rebound} time window ($t_{r}$). 

Solving for the \textit{drop} phase time, $t_{d}$, from Eq. \ref{eq:h0} and \ref{eq:vh0}  and the \textit{rebound} phase time, $t_{r}$, from Eq. \ref{eq:vh1} and \ref{eq:h1}, assuming $[v_{0}, v_{1}, h_{TD}, h_{LO}] = 0$ results in,
\begin{align}
    t_{d} &= \sqrt{\frac{2h_{0}}{g(1 - \gamma_{d} - \gamma_{ld})}}\\
    t_{r} &= \sqrt{\frac{2h_{1}}{g(1 - \gamma_{r} + \gamma_{lr})}}
\end{align}

Using a switching parameters, $\beta$, the two agility metrics, $\nu_{VJA}$ and $\nu_{HA}$, can then be combined  as,
\begin{align}
    \nu &= \frac{h_{1} + \beta{h_{0}}}{t_{s} + t_{r} + \beta t_{d}} \nonumber\\
    &= \frac{h_{1} + \beta{h_{0}}}{t_{s} + \sqrt{\frac{2h_{1}}{g(1 - \gamma_{r} + \gamma_{lr})}} + \beta\sqrt{\frac{2h_{0}}{g(1 - \gamma_{d} - \gamma_{ld})}} }
\end{align}
where, $\beta=0$ gives the vertical jumping agility metric and $\beta=1$ gives the hopping agility metric, thus unifying the two metrics within a mathematical framework to inform future hopping robot design. To include changes in energy during the stance phase ($\zeta_s$), first solve eq. \ref{eq:vh0} and \ref{eq:vh1} for their time and substitute them into eq. \ref{eq:h0} and \ref{eq:h1}, respectively. Solve each of these for their respective velocity at TD ($v_{TD}$) or LO ($v_{LO}$), substitute those into eq. \ref{eq:zeta_s}, and solve for $h_0$ resulting in,
\begin{align}
    h_{0} &= \zeta_{s}^2\frac{h_{1}(1 - \gamma_{r} + \gamma_{lr})}{(1 - \gamma_{d} - \gamma_{ld})} \label{eq:zeta}
\end{align}
where, the previous hop height, $h_0$, and gravitational acceleration, $g$, are represented accordingly. The average \textit{rebound} phase acceleration, $ a_{rebound}=-g(1 - \gamma_{r} + \gamma_{lr})$, and the average \textit{drop} phase acceleration, $a_{drop}=-g(1 - \gamma_{d} - \gamma_{ld})$, are composed of both energy input, $[\gamma_r, \gamma_d]<1$, representing average thrust forces in \textit{rebound} and \textit{drop}, and energy loss, $[\gamma_{lr}, \gamma_{ld}] \geq 0$, representing average drag forces in \textit{rebound} and \textit{drop}; where positive $[\gamma_r, \gamma_d]$ indicates upward thrust and $[\gamma_{lr}, \gamma_{ld}]$ are always opposite the velocity vector. The change in stance phase velocity between TD and LO, $\zeta_s>0$, can capture energy lost, $0<\zeta_s<1$, no energy change, $\zeta_s=1$, and energy gained, $\zeta_s>1$ in the stance phase. Comparing the MultiMo-MHR at differing hopping heights highlights how the hopping height affects the metrics (Table \ref{tab:multimo-agi}); where an increase is observed at lower heights and a saturation appears to occur at higher heights.

Analysis of the metric can highlight key characteristics for enhancing robot agility. First, given Eq. 15, assuming ballistic hopping at a constant height with no inputs or losses, and zero stance time ($[\gamma_{lr}, \gamma_{ld}]=0, \zeta_s=1, [\gamma_r, \gamma_d]=0, h_{1} = h_{0}, t_s=0$), the resulting equation, $\nu=\sqrt{2 g h_1}/2$, shows the metric will increase with increasing gravity or hopping height. Moreover, as the metric switching parameter, $\beta$, cancels out, the two metrics are equivalent under these conditions. 

Second, given Eq. 15, it is clear that increasing the metric requires decreasing the denominator time parameters for the specified heights. The $t_s$ is inversely related to the robot's natural frequency ($\sim \sqrt{m_B/K_s}$); where, the spring constant, $K_s$, and body mass, $m_B$ are represented accordingly. Therefore, the \textit{stance} time ($t_s$) can be minimized by increasing $K_s$ and decrease $m_B$. The \textit{drop} and \textit{rebound} phase times ($t_d,t_r$) can be minimized by maximizing the acceleration magnitudes ($a_{drop}, a_{rebound}$); thus requiring maximum aerial forces. As the MultiMo-MHR can only generate thrust in the positive vertical direction, $0 \leq [\gamma_r, \gamma_d] < 1 $, the metric can be maximized by minimizing $[\gamma_{r}, \gamma_{d}, \gamma_{ld}] = 0$, and maximizing the \textit{rebound} loss $\gamma_{lr}$. This will inherently result in non-constant hopping height unless energy is added in stance; the amount of which can be determined by solving Eq. 16 for the velocity ratio between TD and LO, $\zeta_s$. This does highlight an additional important consideration of the metric. The metric can be increased with a higher previous hop, $h_0$, followed by a lower hop, $h_1$. However, this would reduce the hopping height with each cycle, and therefore could be used to increase agility for individual hops at the expense of long-term agility.

\begin{table}[tbp!]
\caption{MultiMo-MHR Vertical Agility}
\resizebox{\columnwidth}{!}{%
\begin{tabular}{lllllll}
\hline
\multicolumn{1}{|l|}{\textbf{Meas.}} & \multicolumn{1}{l|}{\textbf{Hop}}& \multicolumn{1}{l|}{\textbf{$t_{s}$}}  & \multicolumn{1}{l|}{\textbf{$t_{apogee}$}}& \multicolumn{1}{l|}{\textbf{$t_{cycle}$}}& \multicolumn{1}{l|}{\textbf{$\nu_{VJA}$}} & \multicolumn{1}{l|}{\textbf{$\nu_{HA}$}} \\ 
\multicolumn{1}{|l|}{\textbf{(n=hop num.)}} & \multicolumn{1}{l|}{\textbf{Height(m)}}& \multicolumn{1}{l|}{\textbf{(s)}}  & \multicolumn{1}{l|}{\textbf{(s)}}& \multicolumn{1}{l|}{\textbf{(s)}}& \multicolumn{1}{l|}{\textbf{(m/s) \cite{Haldane2016}}} & \multicolumn{1}{l|}{\textbf{(m/s) \cite{bai_agile_2024}}} \\ \hline
GT (n=16)& 1.03 $\pm$ 0.05& 0.059 $\pm$ 0.001& 0.68 $\pm$ 0.04& 1.18 $\pm$ 0.04&1.52& 1.74\\       
HVSE (n=18)& 1.00 $\pm$ 0.00& 0.058 $\pm$ 0.002& 0.69 $\pm$ 0.06& 1.19 $\pm$ 0.08&1.45& 1.68\\   \hline 
GT (n=19)& 2.02 $\pm$ 0.02& 0.067 $\pm$ 0.001& 0.98 $\pm$ 0.02& 1.68 $\pm$ 0.01&2.05& 2.41\\   
HVSE (n=28)& 1.97 $\pm$ 0.11& 0.07 $\pm$ 0.001& 0.96 $\pm$ 0.11& 1.65 $\pm$ 0.13&2.04& 2.39\\ \hline  
GT (n=16)& 3.03 $\pm$ 0.06& 0.069 $\pm$ 0.001& 1.26 $\pm$ 0.03& 2.14 $\pm$ 0.04&2.40& 2.84\\ 
HVSE (n=39)& 2.85 $\pm$ 0.15& 0.07 $\pm$ 0.001& 1.29 $\pm$ 0.11& 2.13 $\pm$ 0.13&2.20& 2.68\\ \hline 
GT (n=19)& 3.78 $\pm$ 0.13& 0.071 $\pm$ 0.001& 1.71 $\pm$ 0.07& 2.72 $\pm$ 0.08&2.21& 2.78\\ 
HVSE (n=36)& 3.41 $\pm$ 0.29& 0.071 $\pm$ 0.001& 1.64 $\pm$ 0.19& 2.59 $\pm$ 0.24&2.08& 2.63\\ \hline 
GT (max)& 3.92& 0.071& 1.58& 2.60& 2.47& 3.01\\       
HVSE (max)& 4.02& 0.072& 1.85& 2.86&2.18& 2.81\\ \hline
\\ \vspace{-12pt}
\end{tabular}%
}
\begin{tablenotes}
\item Note: Vertical Jumping Agility ($\nu_{VJA}$)\cite{Haldane2016}. Hopping Agility ($\nu_{HA}$) \cite{bai_agile_2024}. 
\end{tablenotes}
\label{tab:multimo-agi}
\end{table}

\begin{table}[tbp]
\caption{Robot Parameters}
\begin{center}
\begin{tabular}{p{0.05\textwidth}p{0.22\textwidth}p{0.08\textwidth}p{0.05\textwidth}}
\hline
\textbf{Parm.} & \textbf{Description} & \textbf{Value} & \textbf{Units} \\[-1ex]
& &  &  \\
\hline
$m_B$ & Body Mass & 0.5619 & kg \\
$m_L$ & Leg Mass & 0.0981 & kg \\
$I_{cmB}$ & Body Rotational Inertia & 0.0012 & kg m$^2$ \\
$I_{cmL}$ & Leg Rotational Inertia & 0.0019 & kg m$^2$ \\
$K_s$ & Main Power Spring Const. & 704 & N/m \\
$K_{lb}$ & Leg-Body Spring Const. & $400 K_s$ & N/m \\
$b_{lb}$ & Leg-Body Damping Coef.& 100 & N s/m \\
$L_1$ & Leg Top to CM$_L$ & 0.1053 & m \\
$L_2$ & Leg Bottom to CM$_L$   & 0.2821 & m \\
$L_3$ & Body CM$_B$ from Top Leg & 0.1191 & m \\
$d_{mtrx}$ & Motor Offset from Body CM & 0.0820 & m \\
$d_{mtry}$ & Motor Offset from Body CM & 0 & m \\
\hline
\end{tabular}
\begin{tablenotes}
\item Full schematic diagram shown in \cite{burns_design_2025}.
\end{tablenotes}
\label{tab:robot_parm}
\end{center}
\end{table}

\section*{Robot Parameters} 
Table \ref{tab:robot_parm} shows the robot parameters used in the dynamic simulation from pervious work \cite{burns_design_2025}.

\section*{Error Metrics}
Five error metrics are shown in Table \ref{tab:error_tabFull_v2} including:
\begin{align}
      M1 &= 1/n_d \sum_j^{n_d} \Bigg( \frac{\frac{1}{n_{p}} \sum_{i}^{n_{p}}  \big| z_{i} - \hat{z}_{i} \big|}{\frac{1}{n_{p}} \sum_{i}^{n_{p}}  \hat{z}_{i}} \Bigg)_j \\
      M2 &= 1/n_d \sum_j^{n_d} \Bigg( \frac{\frac{1}{n_{p}} \sum_{i}^{n_{p}}  \big| \dot{z}_{i} - \hat{\dot{z}}_{i} \big|}{\frac{1}{n_{p}} \sum_{i}^{n_{p}}  | \hat{\dot{z}}_{i} | } \Bigg)_j \\
      M3 &= \frac{1}{n_{HA}} \sum_{i}^{n_{HA}}  \bigg| \frac{z_{HA_i} - \hat{z}_{HA_i}}{\hat{z}_{HA_i}} \bigg| \\
      M4 &= \frac{1}{n_{HA}} \sum_{i}^{n_{HA}}  \big| t_{HA_i} - \hat{t}_{HA_i} \big| \\
      M5 &= \frac{1}{n_{HA}} \sum_{i}^{n_{HA}}  \big| z_{HA_i} - \bar{z}_{HA_i} \big| 
\end{align}
where, $n_d$ is the number of hops, $n_p$ is the number of points per hop (TD to subsequent TD), $n_{HA}$ is the number of hop apexes, and $[z,\hat{z}], [\dot{z},\hat{\dot{z}}], [z_{HA},\hat{z}_{HA}], [t_{HA}, \hat{t}_{HA}]$ represent the estimated and true vertical position, velocity, hop apex height, and hop apex time, accordingly. Finally, $\bar{z}_{HA}$ represents the desired hop apex height.

\begin{table*}[tb!]
\caption{HVSE Performance Metrics}
\begin{center}
\begin{tabular}{lllllllllll}
\cline{4-7}
& \multicolumn{1}{l}{}& \multicolumn{1}{l|}{}& \multicolumn{4}{c|}{\textbf{Set Height (m)}}&\\ \hline
\multicolumn{1}{|c|}{\textbf{Metric}} & \multicolumn{1}{c|}{\textbf{Est.}} & \multicolumn{1}{c|}{\textbf{Meas.}}  & \multicolumn{1}{c|}{\textbf{1}} & \multicolumn{1}{c|}{\textbf{2}} & \multicolumn{1}{c|}{\textbf{3}} & \multicolumn{1}{c|}{\textbf{4}} & \multicolumn{1}{c|}{\textbf{1-4m}}& \multicolumn{1}{c|}{\textbf{Units}} & \multicolumn{1}{c|}{\textbf{Comparison\footnote[3]{} to}} \\ 
\multicolumn{1}{|c|}{\textbf{Description}} & \multicolumn{1}{c|}{Type} & \multicolumn{1}{c|}{}  & \multicolumn{1}{c|}{} & \multicolumn{1}{c|}{} & \multicolumn{1}{c|}{} & \multicolumn{1}{c|}{} & \multicolumn{1}{c|}{}& \multicolumn{1}{c|}{} & \multicolumn{1}{c|}{HVSE}\\ 
\hline
(M1) Mean & BA1& GT& 171.9& 157.7& 193.5& 248.8& 196.8& \%    &t(138) = 30.97, p = 0.00 \\
nMAE Traj Pos.& DR1& GT& 39.10&30.06& 29.10& 74.01& 44.05& \%    &t(138) = 5.62, p = 0.00 \\ 
& KF3& GT& 50.02& 44.12& 51.64& 89.64& 60.45& \% & t(138) = 7.49, p = 0.00\\ 
& HVSE1& GT& 24.64& 20.29& 13.81& 25.60& 21.24 & \% & t(189) = 0.99, p = 0.32   \\
& HVSE2& SE& 25.18& 22.48& 13.88&19.79& 19.46 &\% &Statistical Comparison \\
& HVSE3\textsuperscript{\dag}& GT& 21.36& 19.71& 13.51& 25.38& 20.21& \% &\\
& HVSE4\textsuperscript{\dag}& SE& 25.17& 21.91& 13.48& 19.52& \cellcolor[HTML]{C0C0C0}\textbf{18.97}& \% &\\ 
%
\hline
(M2) Mean & BA1& GT& 143.1& 136.5& 174.7& 245.0& 179.9& \%    &t(138) = 24.15, p = 0.00 \\
nMAE Traj Velo.& DR1& GT& 24.24&19.14& 17.97& 48.46& 28.15& \%    &t(138) = 3.17, p = 0.00 \\ 
& KF3& GT& 30.91& 27.19& 31.70& 57.35& 37.84& \% & t(138) = 5.24, p = 0.00\\ 
& HVSE1& GT& 28.52& 21.52& 12.44& 16.13& 19.58 & \% & t(189) = 0.31, p = 0.76   \\
& HVSE2& SE& 25.53& 21.95& 15.73&17.36& 19.11 &\% &Statistical Comparison \\
& HVSE3\textsuperscript{\dag}& GT& 24.88& 18.78& 11.37& 13.63& 17.08& \% &\\
& HVSE4\textsuperscript{\dag}& SE& 21.98& 19.59& 13.35& 14.78& \cellcolor[HTML]{C0C0C0}\textbf{16.50}& \% &\\ 
\hline
%
(M3)& BA1& GT& 57.34& 51.72& 54.53& 53.23& 53.85& \%    &t(138) = 32.01, p = 0.00 \\
MAPE& DR1& GT& 29.04& 22.13& 22.47& 73.30& 43.36& \%    &t(138) = 5.90, p = 0.00 \\ 
Apex Height& KF3& GT& 37.28& 33.24& 40.75& 81.82& 50.11& \% & t(138) = 7.38, p = 0.00\\ 
& HVSE1& GT& 13.50& 12.25& 8.81& 19.68& 13.77& \%    &t(189) = 0.83, p = 0.41 \\
& HVSE2& SE& 14.79& 14.36& 8.11& 14.62& \cellcolor[HTML]{C0C0C0}\textbf{12.49}& \%    & Statistical Comparison \\ 
\hline
%
(M4)& BA1& GT& 0.3305& 0.4653& 0.6987& 0.9950& 0.7140& s    & t(138) = 16.83, p = 0.00 \\ 
MAE& DR1& GT& 0.0554& 0.0583& 0.0659& 0.2430& 0.1323& s    &t(138) = 2.66, p = 0.00 \\ 
Apex Time& KF3& GT& 0.0726& 0.0830& 0.1178& 0.2483& 0.1388& s & t(138) = 4.17, p = 0.00\\ 
& HVSE1& GT& 0.0611& 0.0581& 0.0369& 0.0956& 0.0641& s    & t(189) = 0.96, p = 0.34\\
& HVSE2& SE& 0.0524& 0.0611& 0.0505& 0.0695& \cellcolor[HTML]{C0C0C0}\textbf{0.0588}& s    & Statistical Comparison \\ 
\hline
(M5) MAE Apex& HVSE1& GT& 0.0477& 0.0294& 0.0562& 0.2190& 0.0912& m    &t(170) = 5.87, p = 0.00 \\
Height (desired)&  HVSE2& SE& 0.0353& 0.0959& 0.1773& 0.5885& \cellcolor[HTML]{C0C0C0}\textbf{0.2597}& m    & Statistical Comparison \\ 
\hline
\\ \vspace{-12pt}
\end{tabular}%
\begin{tablenotes}
\item Error metrics M1-M5 (n=191 hops, Appendix Error Metrics); where \textit{Meas.} column indicates the measurement used for control (true state (GT), estimated state (SE)). \textsuperscript{*} Values indicate the mean of the RMSE value for a complete hop cycle; TD to subsequent TD. \textsuperscript{\dag} Data uses only the aerial phase as detected by the robot for analysis. \footnote[3]{} Comparing the 1-4 m data to the HVSE2 using a two-tailed t-test given the null hypothesis with a 95\% confidence interval. Related works, BA1 \cite{Raibert1984,Yim2020,bai_agile_2024}, KF3 \cite{foxlin_pedestrian_2005,wang_stance-phase_2015,norrdine_step_2016}.
\end{tablenotes}
\label{tab:error_tabFull_v2}
\end{center}
\end{table*}

\section*{Optical Flow Error}
To determine the potential error in the horizontal distance traveled, measured by optical flow. From basic optics, assuming the sensor is traveling parallel to a surface in a single direction, we know the the ratio between the distance an object moves on a camera's image plane, $\hat{d}_c$, divided by the focal distance, $\hat{h}_c$, is equal to the distance the object moves in the world, $d_w$, divided by the distance to the surface, $h_w$, as follows $d_w/h_w = \hat{d}_c/\hat{h}_c$. If the measurement of the distance to the object includes error, $h_{w} = \hat{h}_w+\delta h_w$, then, so too will the distance traveled, $d_w = \hat{d}_w + \delta d_w$. Substituting this into the previous equation results in,
\begin{align}
    \hat{d}_{w} + \delta d_w = (\hat{d}_c/\hat{h}_c) \hat{h}_w (1 + \alpha_h) 
\end{align} 
where, the $\hat{}$ represents the true value, $\delta$ represents the error, and the height error is converted to a percent of the true height, $\delta h_w = \alpha_h \hat{h}_w$. The $\alpha_{h}$ can be approximated as the mean of the normalized mean-absolute-error (nMAE) in the height measurement (Table \ref{tab:error_tabFull_v2}, M1); defined as the mean of the mean-absolute-error in the height divided by the mean of the true height per hop cycle. The error as a percent of the distance traveled can therefore be determined from the ratio of the horizontal distance both with and without the error as,
\begin{align}
    \frac{\hat{d}_{w} + \delta d_w}{\hat{d}_{w}} = 1 + \alpha_{h} \rightarrow \alpha_{h} = \frac{\delta d}{d_{w}}
\end{align} 
where, it can be seen that $\alpha_{h}$ also represents the horizontal error, as a percent of the distance traveled, that would be incurred by replacing an optical flow sensor's range finder with the HVSE.
}

\bibliographystyle{IEEEtran.bst}
\bibliography{IEEEabrv,bibitems_v1}

\newpage

 




\vfill

\end{document}